\documentclass[conference]{IEEEtran}
\IEEEoverridecommandlockouts
\usepackage{cite}
\usepackage{amsmath,amssymb,amsfonts}
\usepackage{algorithmic}
\usepackage[linesnumbered,ruled]{algorithm2e}
\usepackage{graphicx}
\usepackage{textcomp}
\usepackage{xcolor}
\usepackage{booktabs}
\usepackage{multirow}
\usepackage{adjustbox}
\def\BibTeX{{\rm B\kern-.05em{\sc i\kern-.025em b}\kern-.08em
    T\kern-.1667em\lower.7ex\hbox{E}\kern-.125emX}}

\newcommand{\argmin}{arg\,min}
    \makeatletter
    \newcommand{\linebreakand}{%
  \end{@IEEEauthorhalign}
  \hfill\mbox{}\par
  \mbox{}\hfill\begin{@IEEEauthorhalign}
}
\makeatother
\begin{document}

\title{Quality Versus Sparsity in Image Recovery by Dictionary Learning Using Iterative Shrinkage\\
%{\footnotesize \textsuperscript{*}Note: Sub-titles are not captured for https://ieeexplore.ieee.org  and
\thanks{We acknowledge funding by the German Aerospace Center 
(DLR) as part of project AIMS: Artificial Intelligence Meets Space, grant no.\ 50WK2270F, and by the Bundesministerium für Bildung und Forschung (BMBF) within the project \emph{KI@MINT} ("AI-Lab").}
}
\author{\IEEEauthorblockN{1\textsuperscript{st}  Mohammadsadegh Khoshghiaferezaee}
\IEEEauthorblockA{\textit{Institute for Mathematics} \\
\textit{Brandenburg University of Technology}\\
Cottbus, Germany \\
khoshmoh@b-tu.de}
\and
\IEEEauthorblockN{2\textsuperscript{nd} Moritz Krauth}
\IEEEauthorblockA{\textit{Institute for Mathematics} \\
\textit{Brandenburg University of Technology}\\
Cottbus, Germany \\
krautmor@b-tu.de}
\and
\linebreakand
\IEEEauthorblockN{3\textsuperscript{rd} Shima Shabani}
\IEEEauthorblockA{\textit{Institute for Mathematics} \\
\textit{Brandenburg University of Technology}\\
Cottbus, Germany \\
shima.shabani@b-tu.de}
\and
\IEEEauthorblockN{4\textsuperscript{th} Michael Breu\ss{}}
\IEEEauthorblockA{\textit{Institute for Mathematics} \\
\textit{Brandenburg University of Technology}\\
Cottbus, Germany \\
breuss@b-tu.de}
%\and
%\and
}
\maketitle

\begin{abstract}
Sparse dictionary learning (SDL) is a fundamental technique
that is useful for many image processing tasks. 
As an example we consider here image recovery, 
where SDL can be cast as a nonsmooth optimization problem.
For this kind of problems, iterative shrinkage methods
represent a powerful class of algorithms that are subject
of ongoing research. Sparsity is an important property
of the learned solutions, as exactly the sparsity enables 
efficient further processing or storage. The sparsity implies that
a recovered image is determined as a combination of a number of
dictionary elements that is as low as possible.
Therefore, the question arises, to which degree sparsity should 
be enforced in SDL in order to not compromise recovery quality.
In this paper we focus on the sparsity of solutions that
can be obtained using a variety of optimization methods. 
It turns out that there are different sparsity regimes depending on the method in use.
Furthermore, we illustrate that high sparsity does in general not
compromise recovery quality, even if the recovered image is quite 
different from the learning database.
\end{abstract}


\section{Introduction}
To process signals or images, it is considered useful to have models that represent the data. The sparse model offers a particularly effective method, exploiting inherent sparsity and redundancy of signal representations \cite{Ela}. 
It proposes that for a class of signals $\Gamma\subset\mathbb{R}^{m}$, a redundant dictionary $D\in\mathbb{R}^{m\times n}$ ($n\gg m$) exists composed of $n$ prototype signals or atoms.
The core assumption is that any signal in $\Gamma$ can be accurately represented by a sparse linear combination of such atoms. 
Redundancy, meaning that the dictionary's dimensionality exceeds that of the data, facilitates highly sparse representations.

Learning the dictionary from training data rather than relying on pre-defined bases (e.g., wavelets) is a way to emphasize the sparse representation of training signals drawn from $\Gamma$ and is called \emph{sparse dictionary learning} (SDL).
Methods of SDL have been used e.g. for 
Poisson denoising \cite{GL}, in clustering \cite{JNZ}, or deblurring \cite{XMWPZ}.

Turning to SDL optimization, corresponding methods concurrently optimize for an adapted dictionary and its associated sparse representations. This can be formulated as a matrix decomposition problem, in which the training data matrix is approximated as the product of a dictionary matrix and a sparse representation matrix. 
%corresponding to each component of $\Gamma$. 
The solution is obtained through iterative minimization, typically involving alternating phases of \emph{dictionary updating} and \emph{sparse coding}. The implementation of these stages varies across different SDL algorithms \cite{Ela}. 
The question arises, how sensitive high representation quality is with respect to the degree of sparsity enforcement and if this is influenced by the choice of the SDL method. A potentially related issue is, if there is in addition to the obvious sparsity parameter an aspect in the SDL setup that has a large influence on the sparsity property.

In this paper we consider the class of iterative shrinkage methods, which represent a powerful tool for SDL optimization. 
Even though building upon the same concept and solving the same optimization problem, we show that the sparsity of results from different methods may differ in general. We show that there are typical sparsity regimes that do not seem to compromise recovery quality, even generalizing beyond a small set of training images. In addition, we demonstrate that atom dimension has a substantial influence on sparsity.

%In this paper we address this gap in the literature.
%By studying a variety of popular and recent optimization methods for SDL along with varying sparsity parameters, we assess the regimes of sparsity that seem to appear typically in the solutions. 
%In doing this we study in addition a certain range of sparsity parameters of the SDL problem as this is supposedly the most influential factor in the model.

\section{Related Work}
Considering SDL optimization methods, let us briefly mention some of the main developments. The Method of Optimal Directions (MOD) \cite{EAH} focuses on minimizing the representation error while constraining the number of non-zero coefficients in the sparse representation. A probabilistic approach has been proposed in \cite{LS}
building upon independent component analysis and singular value decomposition (SVD). Another approach \cite{LGBB} makes use of a block coordinate relaxation algorithm \cite{SBT} for sparse coding and SVD for atom updates. The K-SVD method \cite{AEB} also utilizes SVD for atom updates and enforces sparsity similar to MOD.
%K-SVD, however, is a batch process requiring the entire training dataset for each iteration.
%, which can be computationally expensive for large or dynamic datasets like video sequences. 

A more flexible setting is presented in terms of an online SDL algorithm in \cite{MBPS}, facilitating especially to learn based on numerous small patches extracted individually from given imagery. 
%It also uses the {\tt LARS} method for sparse recovery and implements block coordinate descent with warm starts in dictionary updating that is parameter-free, without any learning rate tuning.
Employing the online SDL framework, \cite{SMB} studies the impact of training dataset on dictionary construction and recovery quality, providing some insights into method selection and parameter tuning.
%, focusing as in the current paper on the fundamental problem of image recovery. 
% The focus of \cite{SMB} is on a comprehensive computational study comparing state-of-the-art optimization methods, analyzing specifically the impact of the training dataset on dictionary construction. Thereby that analysis 
%includes shrinkage-based methods, such as {\tt FISTA} \cite{BT}, {\tt FPC-BB} \cite{HYZ}, {\tt TwIST} \cite{BF}, {\tt GSCG} \cite{ESM}, and {\tt ISGA} \cite{SB} and 
% explores the influence of implementation parameters on accuracy and computational efficiency.
% We specifically examine in \cite{SMB} how image recovery error changes with increasing dataset size. Our study \cite{SMB} provides insights into method selection and parameter tuning, offering practical guidance for image recovery through online dictionary learning.
However, \cite{SMB} does not consider the sparsity property and related aspects of the setup.

As a perspective conceptually related to the focus of the current paper, one may consider dictionary learning as a variation of sparse matrix factorization \cite{Ela}. Similarly as in our work, in such matrix factorization problems the sparsity of matrices is directly of interest. Since corresponding models may allow the extraction of interpretable patterns from multiway data, there is significant research in developing algorithms that perform constrained low-rank matrix approximations \cite{FR,ZGL}. However, unlike in typical matrix factorization problems, we specifically deal here with SDL for image recovery.

\section{Sparse Dictionary Learning}\label{Acc}
Traditional dictionary learning methods \cite{AEB, OF1} formulate the learning problem as minimizing the reconstruction error on a training dataset $\Gamma\equiv\{p_i\}_{i=1}^{N}$, with $p_i\in \mathbb{R}^{m}$ expressed as:
\begin{equation}
    \label{eq:base_problem}
  \min_{D,\,x_i} \sum_{i = 1}^{N} \lVert  Dx_i-p_i\rVert_2^2  \quad \text{w.r.t.} \quad  \lVert x_i \rVert_0\leq T
\end{equation}
Here, $T$ limits the number of non-zero elements in the sparse representation $x_i\in \mathbb{R}^{n}$, effectively controlling sparsity. The goal is to simultaneously determine the dictionary $D$ and the corresponding sparse representations $x_i$ for the training samples.

Although the $\ell_0$ norm ($\lVert\cdot\rVert_0$) intuitively captures the notion of sparsity by counting non-zero entries, it poses challenges for practical optimization. Therefore, a common approach is to approximate the non-convex $\lVert \cdot\rVert_0$ norm with a convex surrogate. Specifically, in the sparse coding stage, techniques like those in \cite{MBPS} $\ell_1$ norm ($\lVert\cdot\rVert_1$) instead, resulting in the following optimization problem:
\begin{equation}
    \label{eq:base_problem1}
  \min_{D,\,x_i} \frac{1}{N}\sum_{i = 1}^{N}\Big(\frac{1}{2}\lVert Dx_i-p_i \rVert_2^2+\mu\lVert x_i \rVert_1\Big)
\end{equation}
where parameter $\mu$ specifically controls sparsity. 
Let us emphasize, that the $\ell_1$ penalty still encourages sparse solutions as explained geometrically in \cite{Ela}.
The optimization problem (\ref{eq:base_problem1}) exhibits convexity concerning each variable when the other is constant. This allows for an alternating minimization strategy, iteratively optimizing for one variable while fixing the other.

Concerning interpretation of computational results, let us note that we constrain the Euclidean norm of the dictionary atoms.
%to prevent the dictionary $D$ from growing unboundedly, which would lead to trivially small sparse representations $x_i$. 
So, in (\ref{eq:base_problem1}), we restrict $D$ to a convex set $ \mathcal{C}$ defined as, $j=1, \ldots, n$:
\begin{small}
\begin{equation}\label{e:Cdef}
\mathcal{C} = \Big\{D=[d_1, \ldots, d_n]\in\mathbb{R}^{m\times n}|~ d_j\in\mathbb{R}^m,~  
 d_j^Td_j\leq 1\Big\}
\end{equation}
\end{small}

\subsection{Standard Sparse Recovery Methods}

With a fixed dictionary $D$, the sparse recovery stage in \eqref{eq:base_problem1} becomes a convex $\ell_2$-$\ell_1$ optimization problem, commonly known as \textit{basis pursuit denoising}:
\begin{equation}\label{e:defprob1}
 \min_{x}~ \frac{1}{2}\|Dx-p\|_2^2+\mu\|x\|_1 
\end{equation}
%A key benefit of this non-smooth convex problem is the availability of efficient and practical solvers, which accelerate the learning algorithms. 
In this paper, we consider a variety of classical and recent shrinkage-thresholding methods such as {\tt FPC-BB} \cite{HYZ}, {\tt TwIST} \cite{BF}, {\tt GSCG} \cite{ESM}, {\tt ISGA} \cite{SB}, and {\tt smISGA} \cite{SB1}; see e.g. \cite{SMB} for some more details on these methods.

%---------------------------------------------------
%%%%%%%%%%%%%%%%%%SHIMA%%%%%%%%%%%%%%%%%%%%%%%%%%%%%%
\RestyleAlgo{ruled}
\SetKwComment{Comment}{/* }{ */}
\begin{algorithm}
\caption{Dictionary Learning}\label{alg:one}
\KwData{\begin{small}{$D_0\in\mathbb{R}^{m\times n}$ (initial dictionary), $\Gamma$ (patch set), $p\in\mathbb{R}^m$ (image patch), $N$ (number of the patches), $\mu\in\mathbb{R}$ (regularization parameter)}\end{small}}
\KwResult{$D_N$ (learned dictionary)}
\vspace{2mm}
 $\text{List}=\{\tt{FPC-BB, TwIST, GSCG, ISGA, smISGA}\}$\\ 
 Select a {\tt{solver}} from the List\;
 $A_0 \gets 0,~~B_0 \gets 0$ (reset the past information)\;
 \vspace{2mm}
 \For{$k=1$ to $N$}{
  Draw $p_k$ from $\Gamma$\;
  \vspace{2mm}
  \textbf{Sparse Coding Stage:} 
  \vspace{1mm}
    $x_{k} = \underset{x\in\mathbb{R}^n}{\textrm{\argmin}}\frac{1}{2}\|D_{k-1}x-p_k\|_2^2+\mu\|x\|_1$\;  
  $A_k \gets A_{k-1}+x_kx_k^T$\;
  $B_k \gets B_{k-1}+p_kx_k^T$\;
  \vspace{2mm}
  \textbf{Dictionary Updating Stage:}
  \vspace{1mm}
  Update $D_k$ using Algorithm \ref{alg:two}  with $D_{k-1}$ as a warm restart, so that
  %\begin{eqnarray}\label{e:algeq1}
  \begin{small}
  \begin{equation}\label{e:algeq1}
  \begin{split}
    D_k=&\underset{D\in \mathcal{C}}{\textrm{\argmin}}\frac{1}{k}\sum_{i=1}^{k} \frac{1}{2}\|Dx_i-p_i\|_2^2+\mu\|x_i\|_1\\
  =& \underset{D\in \mathcal{C}}{\textrm{\argmin}}\frac{1}{k}\big(\frac{1}{2}\text{Tr}(D^TDA_k)-\text{Tr}(D^TB_k)\big)   
  \end{split}  
  \end{equation}
  \end{small}
 }
\end{algorithm}
%%%%%%%%%%%%%%%%%%%%%%%%%%%%%%%%%%%%%%%%%%%%%%
\RestyleAlgo{ruled}
\SetKwComment{Comment}{/* }{ */}
\begin{algorithm}
\caption{Dictionary Updating}\label{alg:two}
\KwData{$D_{k-1}\in\mathbb{R}^{m\times n}$,\\ $A_k=[a_1,\ldots, a_n]=\sum_{i=1}^{k}x_kx_k^T\in\mathbb{R}^{n\times n}$, $B_k=[b_1,\ldots, b_n]=\sum_{i=1}^{k}p_kx_k^T\in\mathbb{R}^{m\times n}$.}
\vspace{2mm}
\KwResult{$D_k$}
\vspace{2mm}
 \While{convergence}{
 \For{$j=1$ to $n$}{
  Update the $j$-th column to optimize for (\ref{e:algeq1})%\;
  \vspace{2mm}
  \begin{equation}\label{e:algeq2}
  \begin{split}
     u_j&\gets \frac{1}{A_{jj}}\big(b_j-Da_j\big)+d_j\\
  d_j&\gets \frac{1}{\max\big(\|u_j\|_2, 1\big)}u_j 
  \end{split}      
  \end{equation}
 }
 }
\end{algorithm}
%%%%%%%%%%%%%%%%%%%%%%%%%%%%%%%%%%%%%%%%

\subsection{Online Dictionary Learning Algorithm}
Leveraging the online SDL approach outlined in \cite{MBPS,SMB}, we investigate the influence of different sparse recovery techniques. Algorithm \ref{alg:one} summarizes the SDL methodologies.
The algorithm begins by selecting a specific sparse coding solver, which dictates the dictionary learning strategy (lines 1-2). In each iteration, a random patch is extracted from the training dataset $\Gamma$ (line 5). The core of the algorithm (lines 4-10) comprises two essential phases: sparse coding (lines 6) and dictionary atom updating (line 9). To maintain the independence of each iteration, the algorithm resets the accumulated information related to the sparse coefficients computed for previous patches (lines 7-8). This necessitates an initialization step (line 3) to prepare the data structures. In sparse coding stage at iteration $k$, with the dictionary $D_{k-1}$ from the previous iteration held constant, line 6 calculates the sparse representation of the selected patch $p_k$ using the chosen sparse coding solver. In the dictionary updating stage in Algorithm \ref{alg:two} by block coordinate descent method at iteration $k$, the algorithm leverages $D_{k-1}$ as a warm start, incorporating the accumulated and fixed sparse representation of all patches up to that point in $A_k$ and $B_k$. Updates are performed selectively on atoms with non-zero sparse coding contributions ($A_{jj}\neq 0$). It is column-wise, updating one atom at a time while adhering to the constraint $ d_j^T d_j\leq 1$.

%------------------------------------------------
%%%%%%%%%%%%%%%%%%%%%%%%%%%%%%%%%%%%%%%%
\begin{table*}[htb]
\caption{\label{tab1} Reconstruction quality measures, averaged over our test dataset}
{\small
\begin{center}
\begin{tabular}{|c||c|c|c||c|c|c||c|c|c||}
\hline
\textbf{Tested} & \multicolumn{9}{c||}{\textbf{Sparsity Coefficient}} \\
\cline{2-10}
\textbf{Algorithms}
 & \multicolumn{3}{c||}{\textbf{\textit{$2^{-8}$}}}
 & \multicolumn{3}{c||}{\textbf{\textit{$2^{-4}$}}}
 & \multicolumn{3}{c||}{\textbf{\textit{$2^{-1}$}}}\\
\cline{2-10}
& \textbf{RelErr} & \textbf{Dev} & \textbf{SSIM}
& \textbf{RelErr} & \textbf{Dev} & \textbf{SSIM}
& \textbf{RelErr} & \textbf{Dev} & \textbf{SSIM} \\
\hline
{\tt FPC-BB} & 0.00005 & 0.03 & 1.000 & 0.0006 & 0.30 & 0.999 & 0.0039 & 2.01 & 0.999  \\
{\tt TwIST} & 0.00016 & 0.06 & 1.000 & 0.00055 & 0.28 & 1.000 & 0.0049 & 2.13 & 0.999\\
{\tt GSCG}   & 0.00004 & 0.02 & 1.000 & 0.0006 & 0.30 & 0.999 & 0.0042 & 2.17 & 0.999\\
{\tt ISGA}   & 0.00115 & 1.32 & 0.999 & 0.0059 & 6.35 & 0.999 & 0.0178 & 11.23 & 0.994\\
{\tt smISGA} & 0.00078 & 1.07 & 0.999 & 0.0031 & 1.86 & 0.999 & 0.0106 & 6.40 & 0.998\\
\hline
\multicolumn{10}{l}{}
\end{tabular}
\end{center}
}
\end{table*}

%%%%%%%%%%%%%%%%%%%%%%%%%%%%%%%%%%%%%%%%%%%%%%
%----------------------------------------------------------------

\section{Experiments}

%To evaluate our approach, we performed experiments using natural images from the {\tt Kaggle} dataset \cite{Kag}. These experiments were conducted in MATLAB R2024b on an egino BTO system with the following specifications: 128 Intel$^{\circledR}$ Xeon$^{\circledR}$ Gold 6430 processors and 125.1 GiB of RAM. 

%\subsection{Data Preparation}

Let us first elaborate on the experimental setting. The experiments were conducted in MATLAB R2024b on an egino BTO system. To evaluate dictionaries, we considered a colorful random subset of 100 images %with cardinality $100$ and 
of resolution $224\times224$ of the {\tt Kaggle} data set \cite{Kag}.  We did some preprocessing to make them ready for the learning. 

First, we convert them to grayscale and then to a double-precision array. 
If not stated otherwise, we extracted $15000$ random patches as with stride $2$ or $75\%$ overlapping and size $8\times8$ from the random selected images. 
Let us note that subsequently, the patch size and correspondingly the degree
of overlapping may vary. $75\%$ overlapping means each patch has $75\%$ of its area in common with the adjacent patches. So, the process extracts $11881$ patches from each image. Then, patch-wise dictionary learning has been done over $15000$ random patches, constituting our $\Gamma$ set, i.e., $N=15000$ in Algorithm \ref{alg:one}. Based on the patch size ($8\times8$), the size of the produced redundant dictionaries is $64\times256$. 
%\subsection{Important Parameter Settings}

Concerning underlying computational options, the initial dictionary $D_0\in\mathbb{R}^{64\times256}$ is determined by employing a %partial 
discrete cosine transform, the result of which is normalized. 
%We consider two initial point values for sparse recovery as $x_0 = \mathbf{0}_{256 \times 1}$ and
%$x_0=\|D_i^Tp_i\|_{\infty}1_{256\times1}$, where $\lVert\cdot\rVert_{\infty}$ denotes infinity norm of a vector. 
As initial values for sparse recovery we consider $x_0 = \mathbf{0}_{256 \times 1}$.

The stopping condition for all solvers in line 6 of Algorithm \ref{alg:one} is given by \begin{equation}
  \|x_{k+1}-x_{k}\|_2\leq {\epsilon_1}\|x_k\|_2 
\end{equation}
with $\epsilon_1=10^{-7}$ for {\tt GSCG}, {\tt ISGA}, and {\tt smISGA} and $\epsilon_1=10^{-5}$ for {\tt FPC-BB} and {\tt TwIST} or if the number of iterations exceeds 5000000.
For image reconstruction with learned dictionaries in terms of the original image ($\text{{Img}}_{\text{org}}$) and reconstructed image (${\text{Img}}_{\text{rec}}$), the stopping criterion is as
\begin{equation}\label{e:ReErFr0}
  \|\text{{Img}}_{\text{org}}-{\text{Img}}_{\text{rec}}\|_F\leq {\epsilon_2}\|\text{{Img}}_{\text{org}}\|_F 
\end{equation}
using the Frobenius norm, with $\epsilon_2=10^{-10}$ for {\tt GSCG}, {\tt ISGA}, and {\tt smISGA} and $\epsilon_2=10^{-5}$ for {\tt FPC-BB}, and {\tt TwIST}, or if the number of iterations exceeds
5000000. The differences in setting 
$\epsilon_1$ and $\epsilon_2$ are necessary to achieve results of comparable reconstruction quality, see \cite{SMB} for related discussion.
%Again, the difference in $\epsilon_2$ is necessary to achieve results of comparable reconstruction quality.
All other parameters of any given solver are the default as described in the corresponding paper.

Turning to quantitative error measures, we measure the relative error {\tt RelErr} between original and reconstructed images (relative w.r.t. the original image) in the Frobenius norm, as well as the absolute of the maximum deviation {\tt Dev} between original and reconstruction.
Let us note that for these measures, all images are given in terms of standard grey-scale range $[0,255]$, where reconstructed images are not quantized in tonal domain to integer format. Furthermore, we consider the structural similarity index {\tt SSIM} which is a perceptual metric in contrast the beforementioned error measures.

\subsection{Experiment 1: Sparsity vs. Reconstruction Quality}

We now briefly explore if the described SDL approach gives high quality results for image recovery in terms of standard error measures. 
In order to assess quality for different degrees of sparsity, we let the sparsity parameter $\mu$ vary over different orders of magnitude by setting $\mu = 2^{-8}, 2^{-4}, 2^{-1}$ in the optimization problem \eqref{e:defprob1}.
Let us note that $\mu=2^{-8}$ appears to be a standard choice in compressed sensing \cite{ESM,HYZ,SB,SB1}. However, in compressed sensing the underlying problem dimensions are often much larger than in the current work.

%----------------------------------------------
%%%%%%%%%%%%%%%%%%%%%%%%%%%%%%%%%%%%%%%%
\begin{figure}[h]
%\begin{center}
\begin{tabular}{ccc} % 6 columns with spacing in between
    {\includegraphics[width=2.6cm]{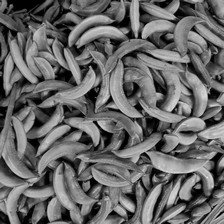}} & {\includegraphics[width=2.6cm]{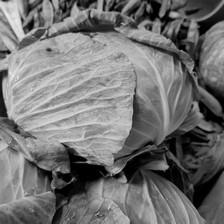}} & {\includegraphics[width=2.6cm]{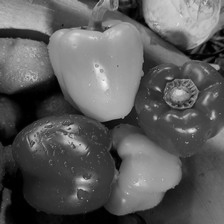}} \\
    {\includegraphics[width=2.6cm]{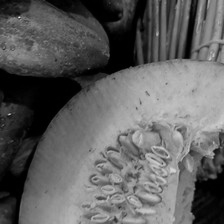}} & {\includegraphics[width=2.6cm]{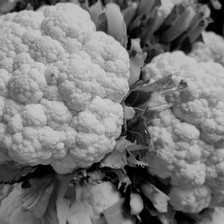}} & {\includegraphics[width=2.6cm]{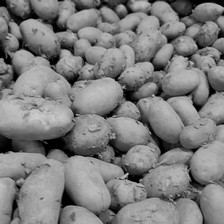}} 
   % {\includegraphics[width=1.5cm]{Patch1.png}} & {\includegraphics[width=1.5cm]{Patch2.png}} & {\includegraphics[width=1.5cm]{Patch3.png}} &\\ 
   % {\includegraphics[width=1.5cm]{Patch4.png}} & {\includegraphics[width=1.5cm]{Patch7.png}} & {\includegraphics[width=1.5cm]{Patch10.png}} \\
\end{tabular}
\caption{\label{Fig:P1}
 Test images from {\tt Kaggle} dataset employed in first two experiments.
}
%\end{center}
\end{figure}
%%%%%%%%%%%%%%%%%%%%%%%%%%%%%%%%%%%%%%%%

Results given in Table \ref{tab1} are averaged over our set of test 
set of six images displayed in Figure \ref{Fig:P1}.
The numbers in Table \ref{tab1} show that even at medium-size sparsity coefficient
$\mu=2^{-4}$, some of the methods give a virtually exact result, as a deviation of less than
$0.5$ grey scale units is supposedly not visible at all. The computationally much more
efficient methods {\tt ISGA} and {\tt smISGA} admit single points of higher deviations,
yet the low relative error indicates that only a small set of pixels is affected at all.
The higher range of sparsity coefficient $\mu=2^{-1}$ means that sparsity is emphasized over
accuracy, so all methods show some deviations in these measures. While deviations of few grey levels in just some pixels as indicated in the table still represent visually convincing reconstructions. This is also the reason why we do not display recovered results here, as they are virtually identical to our test images.

In total, even a higher regime of sparsity parameters $\mu$ does not compromise recovery quality significantly. In turn, the actual degree of sparsity may remain hidden when assessing just quality measures of recovery results.

\subsection{Experiment 2: Sparsity Assessment}

We now consider the sparsity of coefficients in the setting of the first experiment, i.e. for $\mu = 2^{-8}, 2^{-4}, 2^{-1}$ in optimization problem \eqref{e:defprob1}. For sparsity assessment, we employ a binning of the values of the reconstruction coefficients using 100 bins. In this way we obtain histograms of coefficient values. The desirable sparsity property means that a significant peak should be observable for the bins at or around zero, since this would indicate that a corresponding majority of coefficients do not contribute significant image information.

\begin{figure*}[htb]
%\centering{
\begin{center}
\begin{tabular}{ccc}
{\includegraphics[width=4.6cm]{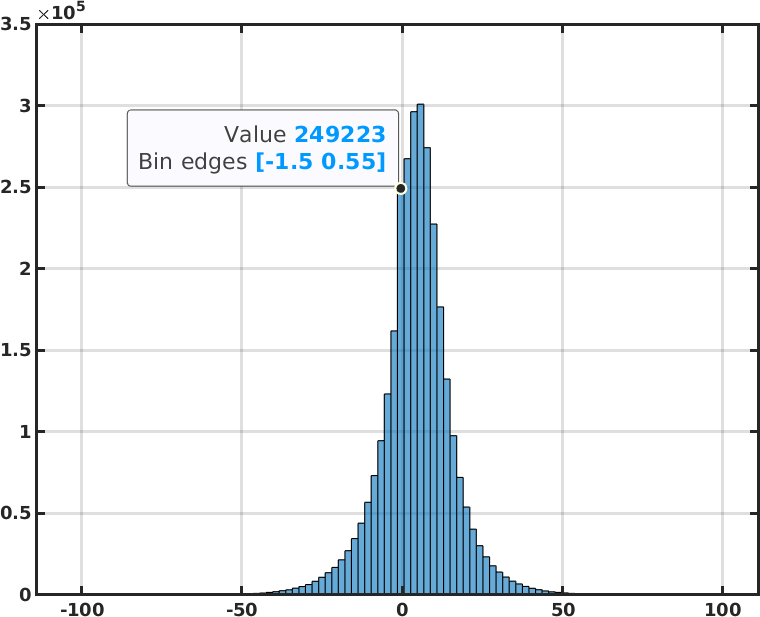}} &
{\includegraphics[width=4.6cm]{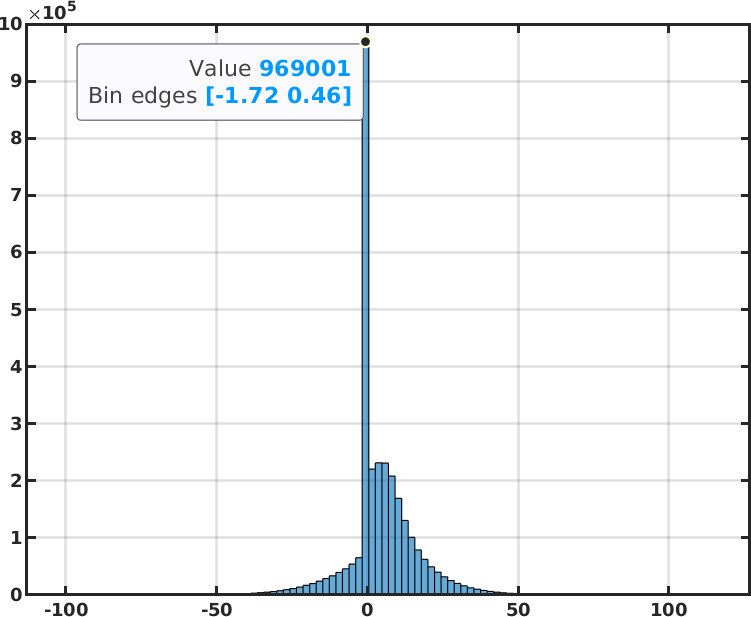}} &
{\includegraphics[width=4.6cm]{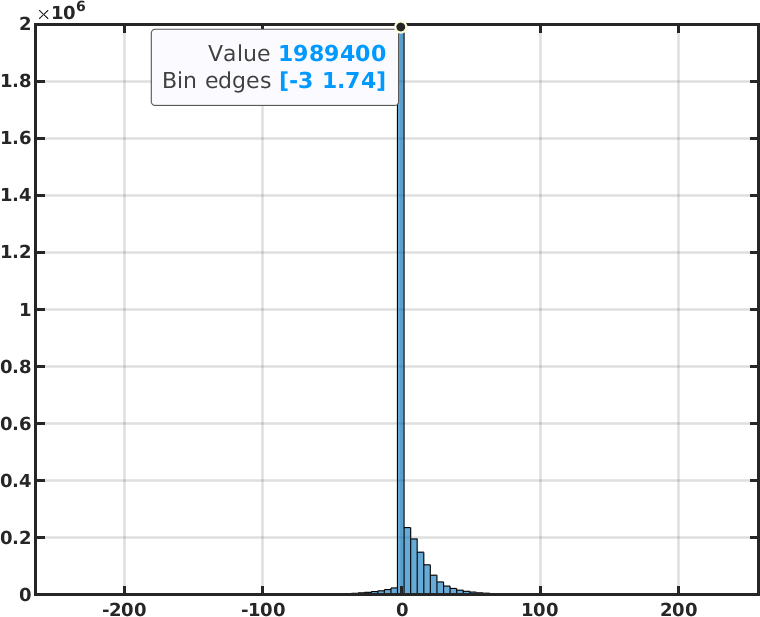}} \\
%{\includegraphics[width=5.2cm]{2TwIST_i2_mu_3.906250e-03_stride_2.png}}\hspace{1mm}{\includegraphics[width=5.2cm]{2TwIST_i2_mu_6.250000e-02_stride_2.png}}\hspace{1mm}{\includegraphics[width=5.2cm]{2TwIST_i2_mu_5.000000e-01_stride_2.png}}\\
%{\includegraphics[width=4.6cm]{2GSCG_i2_mu_3.906250e-03_stride_2.png}} &
%{\includegraphics[width=4.6cm]{2GSCG_i2_mu_6.250000e-02_stride_2.png}} &
%{\includegraphics[width=4.6cm]{2GSCG_i2_mu_5.000000e-01_stride_2.png}} \\
{\includegraphics[width=4.6cm]{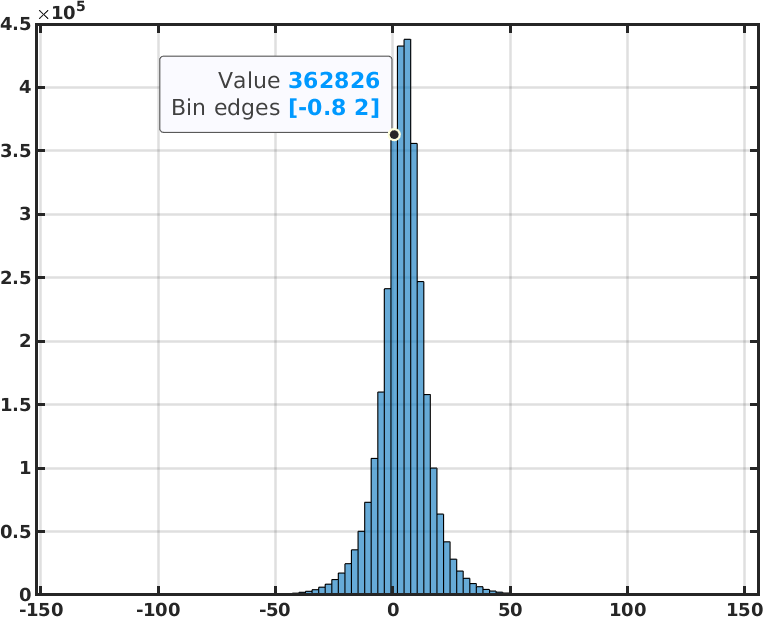}} &
{\includegraphics[width=4.6cm]{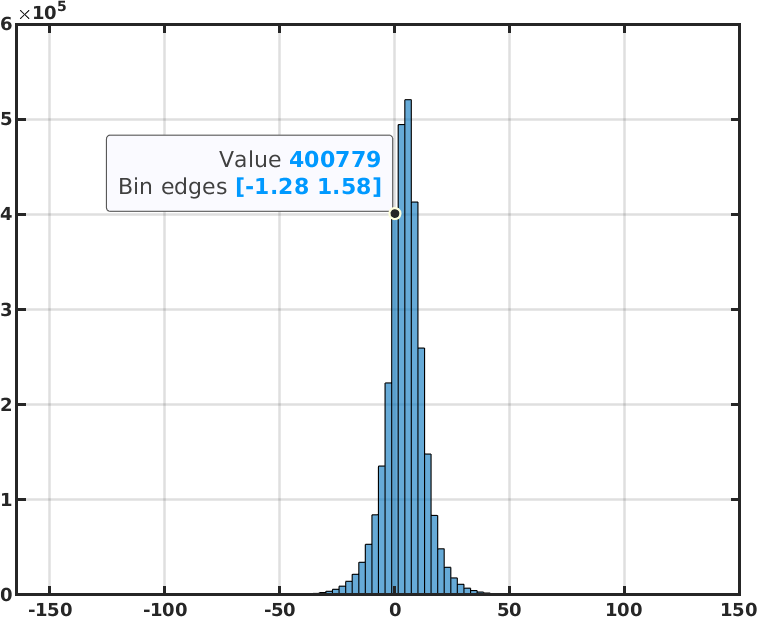}}     &
{\includegraphics[width=4.6cm]{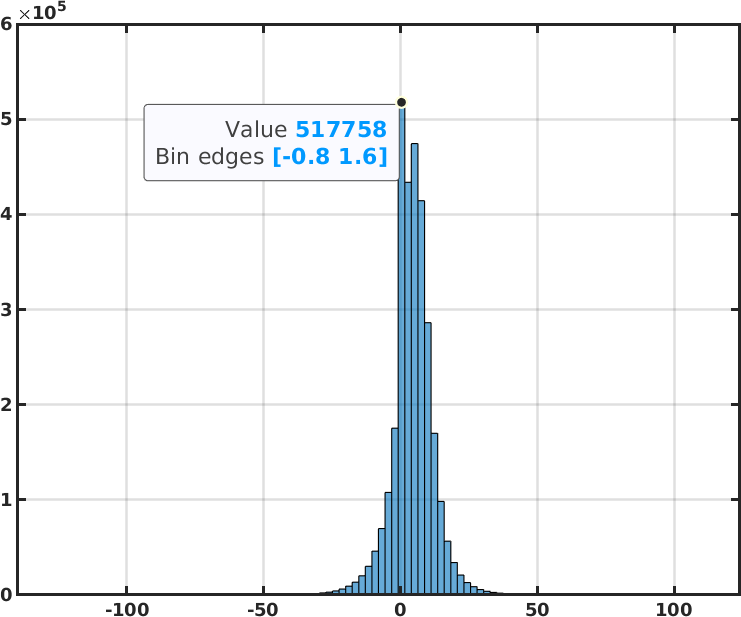}} 
%{\includegraphics[width=5.2cm]{2smISGA_i2_mu_3.906250e-03_stride_2.png}}\hspace{1mm}{\includegraphics[width=5.2cm]{2smISGA_i2_mu_6.250000e-02_stride_2.png}}\hspace{1mm}{\includegraphics[width=5.1cm]{2smISGA_i2_mu_5.000000e-01_stride_2.png}}
\end{tabular}
\caption{\label{Fig:P2}
Sparsity comparison among {\tt FPC-BB}, and {\tt ISGA} (rows) with $\mu=2^{-8},2^{-4},2^{-1}$ (columns) for second test image. 
The results for {\tt TwIST} and {\tt smISGA} are similar to  {\tt FPC-BB} and {\tt ISGA}, respectively. The results of {\tt GSCG} are in between {\tt FPC-BB} and {\tt ISGA}.     
For obtaining similar results showcasing a peak as in the first row with 
{\tt ISGA} and {\tt smISGA}, we have to employ values such as $\mu=2^1,2^2$.
The information in the boxes show the number of values in the peak and the 
corresponding bin edges around zero.
}
\end{center}
\end{figure*}

The main observations of this experiment are summarized at hand of Figure \ref{Fig:P2}. As observable, the default histogram shape appears to be a Gaussian. This is intuitively not surprising, given that in optimization problem (\ref{eq:base_problem1}) the data fidelity term is modeled by employing the $\ell_2$ norm, inducing a Gaussian statistics of coefficients. The role of the sparsity constraint is to extract and diminish from that Gaussian distribution of coefficients the ones that do not hinder reconstruction.

In total, we observe three main sparsity regimes that occur at some stage depending on the optimization method. We conjecture that the results differ depending on the obtained minimization paths for the different methods.

While the low regime of $\mu$ values yields a Gaussian distribution of
coefficients with no visible impact of the sparsity constraint, already the medium regime of $\mu$ values results in significant portion of diminished coefficients for some of the methods, namely {\tt FPC-BB}, {\tt TwIST} and {\tt GSCG}. This is visible at hand of a clear peak, within a Gaussian-type distribution of other coefficients. For {\tt ISGA} and its refined version {\tt smISGA} even higher regimes of $\mu$ values are necessary to obtain such highly sparse solutions. Let us note that, when comparing the widths of the Gaussian part of distributions for all methods and any values of $\mu$, there is no apparent contraction towards zero, meaning that the overall shape is preserved apart from the peak. In highly sparse regime, the peak is highly tweaked and dominates the Gaussian-type distribution of other coefficients.

\begin{figure}[htbp]
%\centering{
\begin{center}
\begin{tabular}{cc}
%{\includegraphics[width=4.2cm]{GSCG-mu_2-to-8-His-Rem.png}} &
%{\includegraphics[width=5.2cm]{FPCBB-mu_2-to-4-His-Rem.png}}%\\
%\hspace{10mm}
%{\includegraphics[width=3.2cm]{GSCG-mu_2-to-8-Diff-Rem.png}} \\
%\hspace{7mm}
{\includegraphics[width=4.2cm]{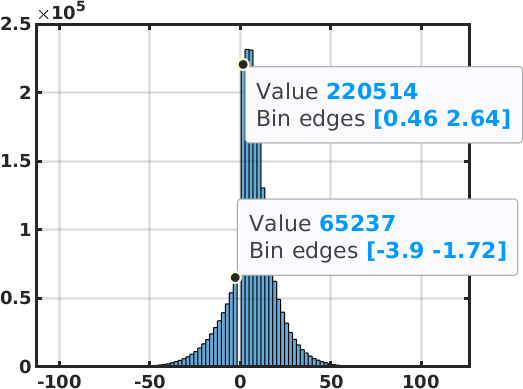}} &
%\\
%\hspace{1mm}
{\includegraphics[width=3.2cm]{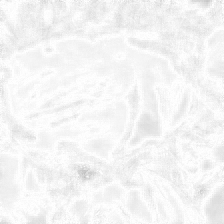}}
\end{tabular}
\caption{\label{Fig:P4}
Histogram after removing peak and corresponding difference images (inverted and scaled by factor 50 for visualization) 
%of {\tt GSCG} in second row - first column of Fig. \ref{Fig:P2} and 
{\tt FPC-BB} in (displayed with peak in first row, second column of Fig. \ref{Fig:P2}). 
}
\end{center}
\end{figure}

As supplementary experiment, we consider reconstructions of second test image when thresholding the peak values, see Figre \ref{Fig:P4}. Plotting the remaining histogram, we clearly see the Gaussian distribution of remaining coefficients. Furthermore, the reconstruction is highly accurate at all tested regimes so that only a significant scaling of differences enables to get a visual impression at all. This observation at hand of Figure \ref{Fig:P4} is quantitatively confirmed by the {\tt SSIM} measures in Table \ref{tab1}.

%-----------------------------------------------

%-----------------------------------------------
%-------------------------------------------------

\subsection{Experiment 3: Influence of Patch Size}

In this experiment we aim to analyze the influence of the patch size respectively the atom size in our SDL setup. Let us note that the total computational setting is of not too high dimensions. For subsequent experiments, our dictionaries are fixed to 512 atoms to properly deal with different patch sizes. The number of rows in turn are given by the atom dimension determined by the patch size. Larger patch sizes also result in higher degree of overlap when keeping the stride 2 fixed as in this experiment.
We combine this experiment by testing at the same time for generalization capability. While the dictionary is learned via {\tt Kaggle} dataset, we now reconstruct the classical {\tt Barbara} test image.

For quantitative evaluation, we propose to fit a Gaussian distribution curve to the arising histograms of coefficients and focus especially on the standard deviations of the Gaussians, as these give a clear indication of the sparsity fitness. In addition, we also consider the previous quality measures.

\begin{figure}[htbp]
%\centering{
\begin{center}
\begin{tabular}{cc}
\includegraphics[width=0.22\textwidth]{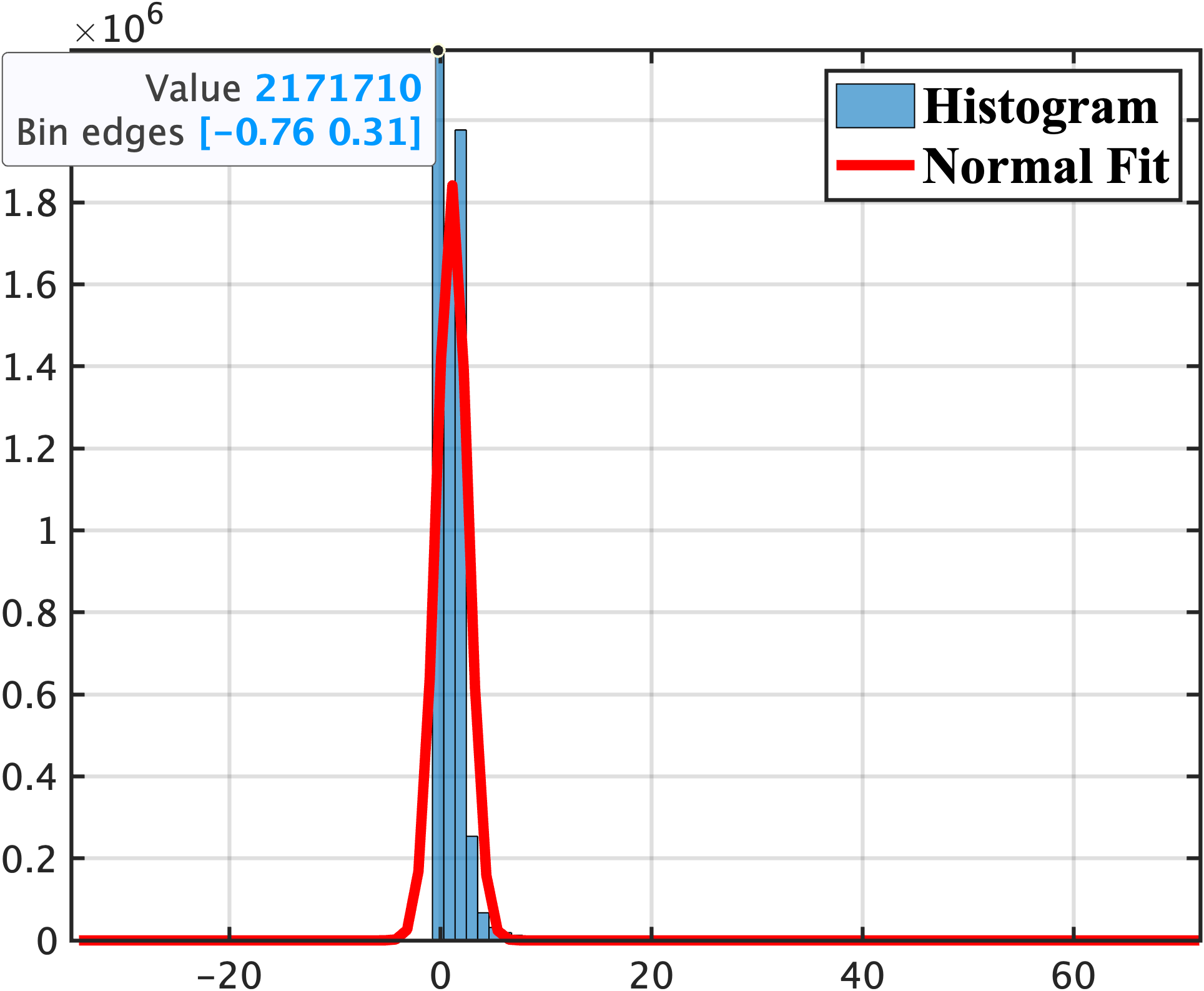} &
\includegraphics[width=0.22\textwidth]{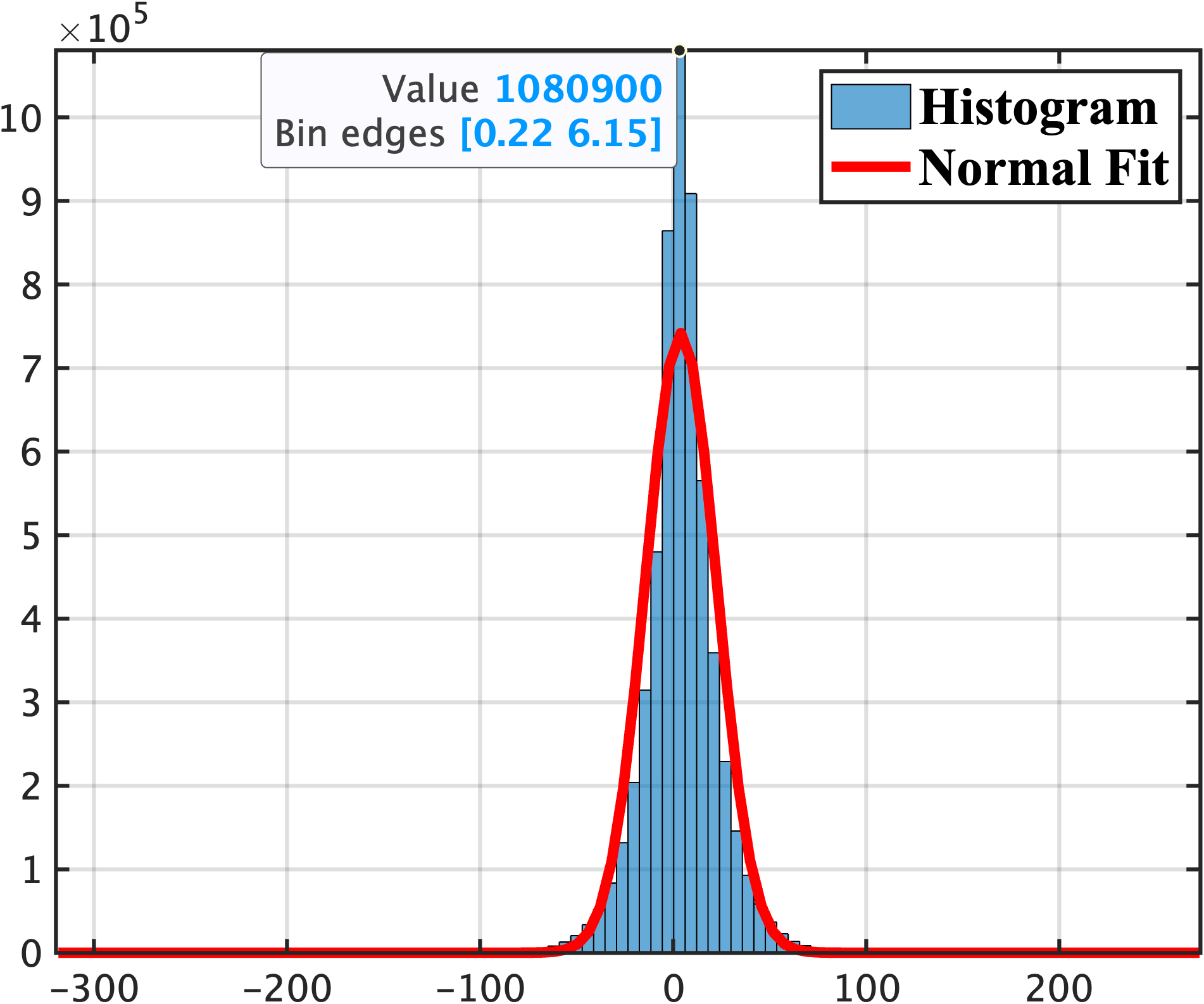} 
\end{tabular}
\caption{\label{Fig:gaussian-fit}
Histogram with Gaussian fit for {\tt FPC-BB} method with patch sizes $4\times4$ and $14\times14$, respectively.
}
\end{center}
\end{figure}
As confirmed by subsequent Tables \ref{barbara:quality-sparse} and \ref{barbara:gaussians} (mean and standard deviation by $\mathrm{m}$ and $\sigma$, respectively), increasing the patch size may lead to a certain loss of quality, but especially the sparsity is degraded. The difference between the two tested algorithms {\tt ISGA} and {\tt FPC-BB} is here, that the {\tt ISGA} method is for $\mu=2^{-4}$ still in the regime of low sparsity and features a clear, unperturbed Gaussian distribution of coefficients. In this regime, the impact of increasing patch size is quite severe in both quality and Gaussian distribution parameters. The {\tt FPC-BB} method is for $\mu=2^{-4}$ in the sparsity regime featuring a clear peak of diminished coefficients. In this sparsity regime we do not observe a significant degradation of quality, but the underlying Gaussian distribution of coefficients as well as the sparsity is clearly affected by increasing patch sizes.

%------------MATIN------------
\begin{table}[h]
{\small
\caption{\label{barbara:quality-sparse}
Atoms=512, $\mu=2^{-4}$, stride =2. Evaluation of Quality: Relative Error, Deviation, and SSIM for Barbara test image $(224\times224)$}
\begin{center}
\begin{tabular}{|c||c|c|c||c|c|c||}
\hline
\textbf{} & \multicolumn{6}{c||}{\textbf{Tested algorithms}} \\
\cline{2-7}
\textbf{Patch}
 & \multicolumn{3}{c||}{\textbf{\textit{\tt ISGA}}}
 & \multicolumn{3}{c||}{\textbf{\textit{\tt FPC-BB}}}\\
\cline{2-7}
& \textbf{RelErr} & \textbf{Dev} & \textbf{SSIM}
& \textbf{RelErr} & \textbf{Dev} & \textbf{SSIM} \\
\hline
4$\times$4 & 0.0003 & 0.291 & 1.00 & 0.0001 & 0.204 & 1.00  \\
% 6x6 (222x222) & 0.00036 & 0.28 & 0.999 & 0.0003 & 0.031 & 1.000  \\
8$\times$8 & 0.0099 & 10.73 & 0.99 & 0.0004 & 0.522 & 0.99  \\
% 10\times10 (220\times220)& 0.0142 & 13.93 & 0.993 & 0.0001 & 0.10 & 1.000 \\
% 12x12 (216x216) & 0.0314 & 30.864 & 0.948 & 0.000062 & 0.41 & 0.999  \\
14$\times$14 & 0.0394 & 43.67 & 0.92  & 0.0006 & 0.646 & 0.99  \\
16$\times$16 & 0.0427 & 44.40 & 0.91 & 0.0005 & 0.546 & 0.99 \\
\hline
\multicolumn{7}{l}{}
\end{tabular}
\end{center}
}
\end{table}
%------------MATOUT-----------
%\FloatBarrier

\begin{table}[htb]
\centering
{\tiny
\caption{\label{barbara:gaussians} Sparsity distribution of {{\tt ISGA}} and {{\tt FPC-BB}} with different $\mu$ values and patch sizes for Barbara test image}
\begin{tabular}{c *{8}{c}}
\toprule
\multirow{3}{*}{\textbf{Patch}} & 
\multicolumn{4}{c}{{{\tt ISGA}}} & 
\multicolumn{4}{c}{{\tt FPC-BB}} \\
\cmidrule(lr){2-5} \cmidrule(lr){6-9}
& \multicolumn{2}{c}{$\mu = 2^{-8}$} & \multicolumn{2}{c}{$\mu = 2^{-4}$} 
& \multicolumn{2}{c}{$\mu = 2^{-8}$} & \multicolumn{2}{c}{$\mu = 2^{-4}$} \\
\cmidrule(lr){2-3} \cmidrule(lr){4-5} \cmidrule(lr){6-7} \cmidrule(lr){8-9}
& $\mathrm{m}$ & $\sigma$ & $\mathrm{m}$ & $\sigma$ & $\mathrm{m}$ & $\sigma$ & $\mathrm{m}$ & $\sigma$ \\
\midrule
4$\times$4  & 1.11 & 1.03 & 1.11 & 1.01 & 1.11 & 1.05 & 1.10 & 1.46 \\
8$\times$8  & 2.25 & 4.27 & 2.25 & 4.25 & 2.24 & 5.50 & 2.24 & 5.64 \\
14$\times$14 & 3.97 & 11.19 & 3.92 & 9.93 & 3.95 & 20.67 & 3.95 & 18.34 \\
16$\times$16 & 4.59 & 11.51 & 4.63 & 11.35 & 4.50 & 28.96 & 4.51 & 26.73 \\
\bottomrule
\end{tabular}
}
\end{table}
%--------------------------------------------

In total, parameter selection in the regime of clearly observable sparsity seems to make SDL more robust towards degradations in quality due to variations in important computational parameters such as the patch size. Taking into account the role of the loss terms in the underlying model (\ref{e:defprob1}), this appears plausible since the sparsity term may be interpreted as a regularizer that encourages to stick to the main atoms responsible for quality recovery.

\section{Conclusion}

Our investigation shows that the regime of the sparsity coefficients appears to be ideally visible as a peak around zero in addition to a Gaussian distribution in the histogram of coefficients. The sparsity regime is not only defined by the sparsity parameter in the SDL optimization model, but depends inherently also on the algorithm in use. The observable image reconstruction quality is in general high by the dictionary learning approach, so that a good reconstruction quality does not indicate a low sparsity quality. In turn, high degree of sparsity seems to make SDL more robust against influence of other parameters defining the computational setup. 

Let us note again that our investigation relies on the use of histograms of coefficients, which appears to be useful in order to identify and classify sparsity regimes that may be favorable for SDL. In the future we aim to build upon this approach for more detailed assessment of SDL robustness as well as automated parameter choices. We also aim to extend our current scope to denoising and other processes of interest.

Learning-based methods like LISTA (Learned ISTA) \cite{GLe} and ADMM-Net (Alternating Direction Method of Multipliers-Net) \cite{SLX} train sparse recovery models with a specific architecture with the aim to produce the best possible approximation of the sparse code and to optimize parameters for the reconstruction task. In future work, we also plan to consider learning models in the sparse coding stage and study possible improvements in accuracy and robustness by learned parameters.


\begin{thebibliography}{00}

\bibitem{AEB}
% Aharon, M., Elad, M., Bruckstein, A.: K-SVD: An algorithm for designing overcomplete dictionaries for sparse representation. IEEE Trans. Signal Process. \textbf{54}(11), 4311--4322 (2006)

M. Aharon, M. Elad, and A. Bruckstein, "K-SVD: An algorithm for designing overcomplete dictionaries for sparse representation," IEEE Trans. Signal Process., vol. 54, no. 11, pp. 4311–4322, 2006.

%\bibitem{BT}
% Beck, A., Teboulle, M.: A fast iterative shrinkage-thresholding algorithm for linear inverse problems. SIAM J. Imaging Sci.
% \textbf{2}(1), 183--202 (2009)

%A. Beck and M. Teboulle, "A fast iterative shrinkage-thresholding algorithm for linear inverse problems," SIAM J. Imaging Sci., vol. 2, no. 1, 2009, pp. 183–202

\bibitem{BF}
% J. M. Bioucas-Dias M. A. T, Figueiredo, ``A new TwIST: Two-step
% iterative shrinkage/thresholding algorithms for image restoration,''
% IEEE Trans. Image Process. vol. 16, 2007, pp. 2992--3004.

J. M. Bioucas-Dias and M. A. T. Figueiredo, "A new TwIST: Two-step iterative shrinkage/thresholding algorithms for image restoration," IEEE Trans. Image Process., vol. 16, no. 12, pp. 2992–3004, 2007.

%\bibitem{CSPPC}
% Chen, Y.C., Sastry, C.S., Patel, V.M., Phillips, P.J., Chellappa, R.: In-plane rotation and scale invariant clustering using dictionaries. IEEE Trans. Image Process. \textbf{22}(6), 2166--2180 (2013)

%Y. C. Chen, C. S. Sastry, V. M. Patel, P. J. Phillips, and R. Chellappa, "In-plane rotation and scale invariant clustering using dictionaries," IEEE Trans. Image Process., vol. 22, no. 6, 2013, pp. 2166–2180.

%\bibitem{Ela}
% Elad, M.: Sparse and Redundant Representation from Theory to
% Application in Signal and Image Processing. Springer, Berlin (2010)

\bibitem{Ela}
M. Elad,``Sparse and redundant representation from theory to
application in signal and image processing,'' Springer, Berlin, 2010.


\bibitem{EA}
% M. Elad, M. Aharon, ``Image denoising via sparse and redundant representations over learned dictionaries,'' IEEE Trans. IP. vol.54(12), pp. 3736–3745, December 2006.

M. Elad and M. Aharon, "Image denoising via sparse and redundant representations over learned dictionaries," IEEE Trans. Image Process., vol. 15, no. 12, pp. 3736–3745, Dec. 2006.

\bibitem{EAH}
% Engan, K., Aase, S.O., Husøy, J.H.: Multi-frame compression: Theory and design. Signal Processing. \textbf{80}(10), 2121--2140 (2000)

K. Engan, S. O. Aase, and J. H. Husøy, "Multi-frame compression: Theory and design," Signal Processing, vol. 80, no. 10, pp. 2121–2140, 2000.

\bibitem{ESM}
% Esmaeili, H., Shabani, S., Kimiaei, M.: A new generalized shrinkage
% conjugate gradient method for sparse recovery. Calcolo
% \textbf{56}(1), 1--38 (2019)

H. Esmaeili, S. Shabani, and M. Kimiaei, "A new generalized shrinkage conjugate gradient method for sparse recovery," Calcolo, vol. 56, no. 1, pp. 1–38, 2019.

%\bibitem{GG}
% Gersho, A., Gray, R.M.: Vector quantization and signal compression. Springer Science \& Business Media (2012). doi.org/10.1007/978-1-4615-3626-0
%A. Gersho and R. M. Gray, Vector Quantization and Signal Compression, Springer Science \& Business Media, 2012. doi.org/10.1007/978-1-4615-3626-0

\bibitem{FR}
S. Foucart and H. Rauhut, ``A mathematical introduction to compressive
sensing,'' Springer, New York 2013.

\bibitem{GLe}
K. Gregor and Y. LeCun, ``Learning fast approximations of sparse coding,'' In Proceedings of the 27th international conference on machine learning, pp. 399-406, June 2010.

\bibitem{GL}
% R. Giryes and M. Elad, ``Sparse coding with an overcomplete Sparsity-based Poisson denoising with dictionary learning,'' IEEE Trans. Image Process. vol.23(12), 2014, pp. 5057--5069.
R. Giryes and M. Elad, "Sparsity Based Poisson Denoising with Dictionary Learning," IEEE Trans. Image Process., vol. 23, no. 12, pp. 5057-5069, December 2014.

\bibitem{HYZ}
% Hale, E.T., Yin, W., Zhang, Y.: Fixed-point continuation applied to compressed sensing: implementation and numerical experiments. J. Comput. Math. \textbf{28} (2), 170--194 (2010)

E. T. Hale, W. Yin, and Y. Zhang, "Fixed-point continuation applied to compressed sensing: implementation and numerical experiments," J. Comput. Math., vol. 28, no. 2, pp. 170–194, 2010. DOI: 10.4208/jcm.2009.10-m1007.

\bibitem{JNZ}
% Jing, L., Ng, M.K., Zeng, T.: Dictionary learning-based subspace structure identification in spectral clustering. IEEE Trans. Neural Netw. and learning systems. \textbf{24}(8), 1188--1199 (2013)
L. Jing, M. K. Ng, and T. Zeng, "Dictionary learning-based subspace structure identification in spectral clustering," IEEE Trans. Neural Netw. Learn. Syst., vol. 24, no. 8, pp. 1188–1199, Aug. 2013.

\bibitem{Kag}
Kaggle dataset. https://www.kaggle.com/

\bibitem{LGBB}
% Lesage, S., Gribonval, R., Bimbot, F., Benaroya, L.: Learning unions of orthonormal bases with thresholded singular value decomposition. InProceedings.(ICASSP'05). IEEE International Conference on Acoustics, Speech, and Signal Processing, (Vol. 5, pp. v-293). (2005)

S. Lesage, R. Gribonval, F. Bimbot, and L. Benaroya, "Learning unions of orthonormal bases with thresholded singular value decomposition," Proc. IEEE Int. Conf. Acoust., Speech, Signal Process. (ICASSP), vol. 5, pp. 293–296, 2005.

\bibitem{LS}
% Lewicki, M.S., Sejnowski, T.J.: Learning overcomplete representations. Neural Computation. \textbf{12}(2),337--365 (2000)

M. S. Lewicki and T. J. Sejnowski, "Learning overcomplete representations," Neural Computation, vol. 12, no. 2, pp. 337-365, 2000.
\bibitem{MBPS}
% Mairal, J., Bach, F., Ponce, J., Sapiro, G.: Online dictionary learning for sparse coding. In: Proceedings of the 26th annual international conference on machine learning. 689--696 (2009)

J. Mairal, F. Bach, J. Ponce, and G. Sapiro, "Online dictionary learning for sparse coding," in Proceedings of the 26th Annual International Conference on Machine Learning (ICML), pp. 689–696, 2009.

\bibitem{MMYZ}
% Ma, L., Moisan, L., Yu, J., Zeng, T.: A dictionary learning approach for Poisson image deblurring. IEEE Trans. Med. Imag. \textbf{32}(7), 1277--1289 (2013)

L. Ma, L. Moisan, J. Yu, and T. Zeng, "A dictionary learning approach for Poisson image deblurring," IEEE Trans. Med. Imaging, vol. 32, no. 7, pp. 1277–1289, July 2013.

\bibitem{OF1}
% B.A. Olshausen and D.J. Field, ``Sparse coding with an overcomplete basis set: A strategy employed by V1?,'' Vision research. vol.37(23), 1997, pp. 3311--3325.

B. A. Olshausen and D. J. Field, "Sparse coding with an overcomplete basis set: A strategy employed by V1?," Vision Research, vol. 37, no. 23, pp. 3311–3325, Dec. 1997.

%\bibitem{RB}
% S. Roth, and M.J. Black, ``Fields of experts: A framework for learning image priors,'' In 2005 IEEE Computer Society Conference on Computer Vision and Pattern Recognition (CVPR'05), vol. 2, pp. 860--867. IEEE, 2005.

%S. Roth and M. J. Black, "Fields of experts: A framework for learning image priors," in Proc. IEEE Conf. Comput. Vis. Pattern Recognit. (CVPR), vol. 2, pp. 860–867, 2005.

\bibitem{SBT}
% Sardy S, Bruce AG, Tseng P. Block coordinate relaxation methods for nonparametric wavelet denoising. Journal of computational and graphical statistics. \textbf{9}(2), 361--379 (2000)

S. Sardy, A. G. Bruce, and P. Tseng, "Block coordinate relaxation methods for nonparametric wavelet denoising," J. Comput. Graph. Stat., vol. 9, no. 2, pp. 361-379, 2000.


\bibitem{SB} 
% S. Shabani and M. Breu{\ss},
% ``An efficient line search for sparse reconstruction,'' In 9th international conference on scale space and variational methods in computer vision (SSVM), LNCS, vol. 14009, 2023, pp. 471--483.
% Springer, Cham 2023.

S. Shabani and M. Breu{\ss}, "An efficient line search for sparse reconstruction," in 9th International Conference on Scale Space and Variational Methods in Computer Vision (SSVM), LNCS, vol. 14009, Springer, Cham, pp. 471–483, 2023.

\bibitem{SB1} 
% S. Shabani and M. Breu{\ss},
% ``Semi-monotone Goldstein line search strategy with application in sparse recovery,'' In 10th international conference on scale space and variational methods in computer vision (SSVM), 2025, peer-reviewed to appear in LNCS  Springer series.

S. Shabani and M. Breu{\ss}, "Semi-monotone Goldstein line search strategy with application in sparse recovery," in Proc. 10th
International Conference on Scale Space and Variational Methods in Computer Vision (SSVM), LNCS, vol. 15668, Springer, Cham, pp. 69–81, 2025.

\bibitem{SMB} 
% S. Shabani, M. Khoshghiaferezaee, and M. Breu{\ss},
% ``Sparse dictionary learning for image recovery by iterative shrinkage,'' In IntelliSys 2025, peer-reviewed, to appear in LNNS  Springer series.
S. Shabani, M. Khoshghiaferezaee, and M. Breu{\ss}, "Sparse dictionary learning for image recovery by iterative shrinkage," in IntelliSys 2025, to appear in Lecture Notes in Networks and Systems (LNNS), Springer. https://arxiv.org/abs/2503.10732

%\bibitem{WF}
% Y. Weiss and W.T. Freeman, ``What makes a good model of natural images?,'' In 2007 IEEE conference on computer vision and pattern recognition (CVPR), pp. 1--8. IEEE, 2007.
%Y. Weiss and W. T. Freeman, "What makes a good model of natural images?," in Proc. 2007 IEEE Conference on Computer Vision and Pattern Recognition (CVPR), pp. 1–8, 2007.

\bibitem{SLX}

J. Sun, H. Li, and Z. Xu, ``Deep ADMM-Net for compressive sensing MRI,'' Advances in neural information processing systems, 29, 2016.

\bibitem{XMWPZ}
% Xiang, S., Meng, G., Wang, Y., Pan, C., Zhang, C.: Image deblurring with coupled dictionary learning. International Journal of Computer Vision. \textbf{114},248--271 (2015)
S. Xiang, G. Meng, Y. Wang, C. Pan, and C. Zhang, "Image deblurring with coupled dictionary learning," International Journal of Computer Vision, vol. 114, pp. 248–271, 2015.

\bibitem{ZGL}
T. Zhang, B. Ghanem, S. Liu, C. Xu, and N. Ahuja, ``Low-rank sparse coding for image classification,'' In Proceedings of the IEEE international conference on computer vision, pp. 281-288, 2013.
%%%%%%%%%%%%%%%%%%%%%%%%%%%%%%%%%%%%%%%%%%%%%%%%%%%%%%%%%%%%
%--------------------------------------------
\end{thebibliography}
\end{document}